\documentclass{article}

\usepackage{arxiv}

\usepackage[utf8]{inputenc} 
\usepackage[T1]{fontenc}    
\usepackage[hypertexnames=false]{hyperref}       
\usepackage{bookmark}
\usepackage{xurl}           
\usepackage{url}            
\usepackage{amsfonts}       
\usepackage{nicefrac}       
\usepackage{microtype}      
\usepackage{graphicx}
\usepackage{amsmath}
\usepackage[numbers]{natbib}

\usepackage{booktabs}             
\usepackage{multirow}             
\usepackage{makecell}             
\usepackage{tabularx}             

\usepackage[table]{xcolor} 
\usepackage{booktabs}      
\usepackage{multirow}      
\usepackage{algorithm}
\usepackage{algorithmicx}
\usepackage{algpseudocode}
\usepackage{listings}

\usepackage{xcolor}
\usepackage{float}
\usepackage{enumitem}
\usepackage{placeins}

\usepackage{tikz}
\usepackage{pgfplots}
\pgfplotsset{compat=1.18} 

\makeatletter
\providecommand*{\toclevel@algorithm}{0}
\makeatother

\graphicspath{ {./images/} }

\title{TriSP: Tri-Signal Structured Pruning for Large Language Models}

\author{
 Manel Kara laouar\thanks{Corresponding author}  \\
National School of Artificial Intelligence (ENSIA) \\
Sidi Abdellah Campus,  Algiers, Algeria\\
  \texttt{manel.karalaouar@ensia.edu.dz} \\
   \And
  Soumia Bouyahiaoui \\
  National School of Artificial Intelligence (ENSIA) \\
  Sidi Abdellah Campus,  Algiers, Algeria\\
\texttt{soumia.bouyahiaoui@ensia.edu.dz} \\
  \And
 Aicha Boutorh \thanks{Corresponding author}  \\
  National School of Artificial Intelligence (ENSIA) \\
  Sidi Abdellah Campus,  Algiers, Algeria\\
  \texttt{aicha.boutorh@ensia.edu.dz} \\ 
}

\begin{document}
\maketitle
\begin{abstract}

Large language models (LLMs) achieve strong performance across diverse tasks but their deployment is constrained by the memory and compute cost of their parameters. Structured pruning addresses this by removing entire structures such as attention heads and Multi-Layer Perceptron (MLP) neurons to produce smaller dense models that run efficiently on standard hardware. However, existing methods rely on either gradient-based importance estimation, which is memory-prohibitive, or activation-based statistical proxies, which do not directly measure the effect of removal on the loss. Furthermore, the interaction between the importance criterion and the post-pruning recovery strategy has not been systematically studied. We propose TriSP (Tri-Signal Structured Pruning), an importance metric that combines weight magnitude scaled by activation norm with first-order gradient sensitivity via a geometric mean, producing a channel-level score that captures both structural and loss-sensitivity signals. Combined with adaptive per-layer budget allocation and low-rank adaptation (LoRA) recovery, TriSP achieves the lowest perplexity and highest zero-shot accuracy across all tested configurations, reaching 6.80 WikiText-2 perplexity at 20\% pruning on LLaMA-7B. Inference throughput improves by 82\% at 50\% pruning, while still maintaining competitive performance.
\end{abstract}



\section{Introduction}

Large language models (LLMs) based on the transformer architecture~\citep{ashish2017attention} have become the dominant paradigm for natural language processing. Models such as GPT-4~\citep{achiam2023gpt}, LLaMA~\citep{touvron2023llama}, Mistral~\citep{jiang2023mistral}, and DeepSeek~\citep{bi2024deepseek} demonstrate strong performance across several tasks. However, their practical deployment is constrained by the computational and memory resources they require: a 7-billion parameter model in half-precision occupies approximately 13\,GB of GPU memory and requires hundreds of billions of multiply-accumulate operations per forward pass~\citep{zhu2024survey}.

\emph{Structured pruning} reduces this cost by removing entire architectural components: attention heads, Multi-Layer Perceptron (MLP) neurons, or channels, producing smaller dense models that benefit from standard GPU matrix operations. Unlike unstructured pruning, which zeros individual weights and requires specialized sparse kernels for speedup, structured pruning yields direct latency and memory reductions on commodity hardware. The central challenge is identifying which structures to remove with minimal quality degradation, which requires both an accurate importance criterion and an allocation strategy that distributes the pruning budget across layers in proportion to their redundancy.

Existing approaches address these requirements with complementary but incomplete solutions. FLAP~\citep{an2024fluctuation} uses activation variance scaled by squared weight magnitude, a statistical proxy that is fast to compute but does not directly measure the effect of removal on the model's loss. LLM-Pruner~\citep{ma2023llm} uses Taylor-based gradient importance with dependency graph analysis, which more directly estimates pruning impact but requires storing gradient tensors for all pretrained weights, making it memory-prohibitive for large models. Both methods apply post-pruning recovery (bias compensation for FLAP, LoRA fine-tuning for LLM-Pruner) to restore quality, but no prior work has evaluated how the choice of recovery strategy affects the final quality.

Two research gaps motivate this work. First, gradient augmentation has consistently improved unstructured pruning: GBLM-Pruner~\citep{das2023beyond} adds a gradient term to Wanda's~\citep{sun2023simple} activation-norm criterion, and Wanda++~\citep{yang2025wanda++} extends this with regional gradients. However, this approach has not been transferred to structured pruning, where gradient-based and activation-based methods remain separate. Second, FLAP's bias compensation is presented as sufficient to eliminate fine-tuning, yet quality degrades substantially at aggressive ratios, and no prior work has compared this static correction against LoRA recovery~\citep{hu2022lora} under the same pruning criterion. This paper makes two contributions:

\begin{enumerate}
    \item \textbf{TriSP (Tri-Signal Structured Pruning)}, an importance metric that combines activation-weighted magnitude with gradient sensitivity via a geometric mean, transferring gradient augmentation from unstructured to structured pruning.
    \item A \textbf{systematic comparison} of bias compensation vs.\ LoRA recovery under the same pruning criterion, and comprehensive evaluation across five models.
\end{enumerate}

The remainder of this paper is organized as follows: Section~\ref{sec:related} reviews related work, Section~\ref{sec:method} describes the proposed methodology, Section~\ref{sec:experiments} presents the experimental setup and results, and Section~\ref{sec:conclusion} concludes the paper.

\section{Related Work}
\label{sec:related}

Structured pruning for LLMs removes entire structural units (heads, neurons, channels) from a pretrained model to produce a smaller dense architecture that directly benefits from standard GPU kernels~\citep{he2023structured}. Methods differ along two axes: the importance criterion used to rank structures and whether post-pruning recovery is applied.

\subsection{Gradient-Based Methods}

First-order Taylor expansion provides a principled estimate of how much removing a structural unit affects the loss. LLM-Pruner~\citep{ma2023llm} applies this criterion with approximate Hessian information, coupling parameters into structural groups (heads, MLP neurons) via a dependency graph and recovering quality with a lightweight LoRA adapter. However, it requires storing full gradient tensors for all pretrained weights, making it memory-intensive at scale. Two subsequent methods address this bottleneck by computing Taylor importance from the small LoRA adapter matrices instead: LoRAPrune~\citep{zhang2024loraprune} enables pruning of LLaMA-65B on a single A100 GPU and applies iterative progressive pruning, while DyLoRA-Prune~\citep{li2024dynamic} replaces fixed-rank LoRA with Dynamic Low-Rank Adaptation, learning representations across a range of ranks simultaneously so that lower ranks capture the most critical information.

An alternative line of work uses second-order information. SlimGPT~\citep{ling2024slimgpt} extends the Optimal Brain Surgeon (OBS) framework~\citep{hassibi1992second}, previously applied to unstructured pruning (SparseGPT~\citep{frantar2023sparsegpt}) and quantization (GPTQ~\citep{frantar2022gptq}), introducing Batched Greedy Pruning with grouped Cholesky decomposition for heads and dynamic group sizing for MLP channels. Even without fine-tuning, it reaches 52.23\% zero-shot average at 50\% on LLaMA-7B, outperforming LLM-Pruner with LoRA, while requiring only 7~GB of GPU memory at 20\%.

\subsection{Gradient-Free Methods}

Activation-based criteria avoid gradient computation entirely. FLAP~\citep{an2024fluctuation} scores channels by the sample variance of input activations weighted by squared weight-column norm, replacing low-fluctuation channels with a fixed bias correction that requires no gradients. Similarly, Bonsai~\citep{kolawole2024everybody} frames importance as a regression problem, fitting a linear model over randomly sampled sub-models to estimate each module's contribution globally, enabling pruning on a single 48~GB A6000 where gradient-based methods require 80--640~GB. Probe Pruning~\citep{le2025probe} extends the activation-based paradigm by making decisions online: a lightweight probe subset (5\% of samples) runs ahead through a few layers, fusing resulting states with historical calibration states so that current batch characteristics dominate when they deviate from calibration, surpassing FLAP (61.0\% vs.\ 60.6\% commonsense accuracy at 40\% on LLaMA-2-13B) with only 1.5\% overhead in FLOPs.

A separate family of gradient-free methods operates at coarser granularity. SliceGPT~\citep{ashkboos2024slicegpt} exploits the computational invariance of RMSNorm transformers under orthogonal rotations, concentrating signal into leading PCA components and slicing off small-eigenvalue dimensions uniformly across the model. At the layer level, ShortGPT~\citep{men2025shortgpt} removes entire transformer layers based on Block Influence, observing that middle layers carry the most redundancy, while BlockPruner~\citep{zhong2025blockpruner} refines this by treating MHA and MLP blocks as separate pruning units and finding asymmetric redundancy: below 17\% pruning MHA blocks are more dispensable, but beyond that threshold MHA becomes critical while MLP remains tolerant.

\subsection{Gradient Augmentation in Unstructured Pruning}

In unstructured pruning, Wanda~\citep{sun2023simple} demonstrates that a simple activation-aware criterion $S_{ij} = |W_{ij}| \cdot \|X_j\|_2$ can match SparseGPT's accuracy without weight reconstruction. GBLM-Pruner~\citep{das2023beyond} extends this score with a gradient term, consistently outperforming both Wanda and SparseGPT on LLaMA-2-7B at 50\% sparsity. Wanda++~\citep{yang2025wanda++} further reduces memory cost by restricting gradient computation to single decoder blocks and applies regional weight updates to minimize pruning-induced output discrepancy. However, this gradient augmentation has not been transferred to structured channel scoring. 

In this work, we propose TriSP, which bridges this gap by combining activation-weighted magnitude with gradient sensitivity for structured pruning.

\section{Methodology}
\label{sec:method}

\begin{figure}[h]
    \centering
    \includegraphics[width=\textwidth]{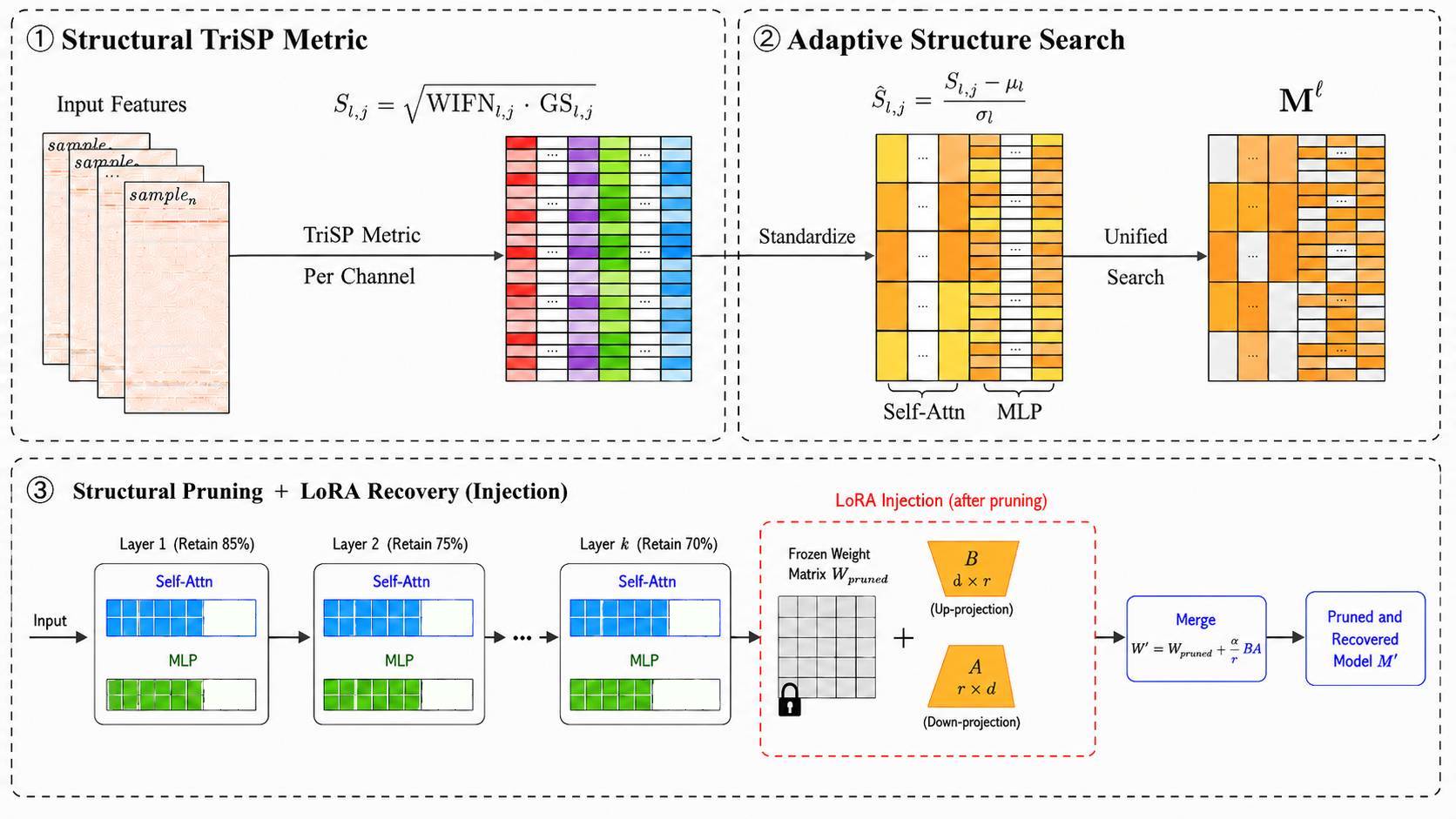}
       \caption{Overview of the proposed TriSP structured pruning pipeline.
             \textbf{(1)}~Compute tri-signal importance scores $S_{l,j}$, the geometric mean of WIFN$_{l,j}$ and GS$_{l,j}$, combining weight magnitude, gradient sensitivity, and activation norm per channel from calibration data;
    \textbf{(2)}~Standardize scores across layers via z-score normalization and apply a unified threshold;
    \textbf{(3)}~Physically remove pruned heads and neurons with adaptive per-layer retention ratios, then inject LoRA adapters into the frozen pruned weights and merge after fine-tuning.}
    \label{fig:pipeline}
\end{figure}

The proposed method operates as a three-stage pipeline (Figure~\ref{fig:pipeline}): importance estimation, structural pruning via adaptive layer-module (AL-AM) allocation, and an optional low-rank adaptation (LoRA) recovery.

============================================================

\subsection{Calibration}

Importance estimation requires collecting activation statistics and gradients from a small calibration set. For each sample, we perform a full forward pass through the model to obtain logits, compute the per-sample cross-entropy loss $\mathcal{L}$ on next-token prediction:
\begin{equation}
    \mathcal{L} = -\frac{1}{T-1} \sum_{t=1}^{T-1}
    \log P(x_{t+1} \mid x_1, \dots, x_t)
    \label{eq:ce_loss}
\end{equation}
where $T$ is the sequence length and $x_t$ is the token at position $t$. We then execute a backward pass to produce weight gradients. The absolute gradients are accumulated across all $N$ calibration samples to form a per-weight average gradient magnitude:

\begin{equation}
\bar{G}_{l,ij} = \frac{1}{N} \sum_{t=1}^{N}
\left| \frac{\partial \mathcal{L}^{(t)}}{\partial W_{l,ij}} \right|
    \label{eq:grad_avg}
\end{equation}
where $W_{l,ij}$ denotes the weight at row $i$, column $j$ of a linear layer in transformer layer $l$.

During the forward pass, input activation statistics are collected at the output projection and down projection of each transformer layer as a running squared $\ell_2$ norm:

\begin{equation}
    a_{l,j}^{(t)} = \frac{n}{n + b} \, a_{l,j}^{(t-1)}
    + \frac{\|X_{l,j}^{(t)}\|_2^2}{n + b}
    \label{eq:running_norm}
\end{equation}

where $a_{l,j}^{(t)}$ is the accumulated squared activation norm for input channel $j$ in layer $l$ at step $t$, $X_{l,j}^{(t)}$ is the corresponding activation vector, $n$ is the number of samples seen so far, and $b$ is the current batch size. After all calibration batches, $a_{l,j}$ converges to the mean squared activation norm at channel $j$.

Although TriSP requires an additional backward pass compared to FLAP, this roughly doubles the per-sample compute only during calibration, which is a one-time cost that completes in minutes on a single GPU, and the improved pruning quality justifies the overhead.

\subsection{Tri-Signal Structured Metric}

We propose a tri-signal importance metric that combines a structural signal from FLAP~\citep{an2024fluctuation} with a gradient-based sensitivity signal. For a linear layer with weight matrix $W \in \mathbb{R}^{d_{\text{out}} \times d_{\text{in}}}$, we define two per-channel components. The first is the \emph{Weighted Input Feature Norm} (WIFN), which captures weight magnitude scaled by activation norm:
\begin{equation}
    \text{WIFN}_{l,j} = \frac{1}{d_{\text{out}}}
    \sum_{i=1}^{d_{\text{out}}} |W_{l,ij}|
    \cdot \sqrt{a_{l,j} + \epsilon}
    \label{eq:wifn}
\end{equation}
where $W_{l,ij}$ is the weight connecting input channel $j$ to output channel $i$ in layer $l$, $d_{\text{out}}$ is the number of output channels, $a_{l,j}$ is the accumulated squared activation norm from Equation~\ref{eq:running_norm}, and $\epsilon = 10^{-6}$.

The second is the \emph{Gradient Sensitivity} (GS) term, which captures how much the loss depends on each channel, again scaled by activation norm:
\begin{equation}
    \text{GS}_{l,j} = \left( \frac{1}{d_{\text{out}}}
    \sum_{i=1}^{d_{\text{out}}} |W_{l,ij}| \cdot \bar{G}_{l,ij} \right)
    \cdot \sqrt{a_{l,j} + \epsilon}
    \label{eq:gs}
\end{equation}
where $\bar{G}_{l,ij}$ is the averaged absolute gradient from Equation~\ref{eq:grad_avg}.
The final importance score is the geometric mean of these two components:
\begin{equation}
    S_{l,j} = \sqrt{\text{WIFN}_{l,j} \cdot \text{GS}_{l,j}}
    \label{eq:gw}
\end{equation}

This formulation captures three complementary signals:
\begin{itemize}
    \item \textbf{Weight magnitude} $|W_{l,ij}|$: channels connected through large weights carry more information.
    \item \textbf{Gradient magnitude} $\bar{G}_{l,ij}$: channels with high gradient magnitude indicate higher importance for model performance.
    \item \textbf{Input activation norm} $\sqrt{a_{l,j}}$: channels that receive large activations during calibration contribute more to the layer output, consistent with the observation by \citet{sun2023simple} that outlier features in LLMs carry disproportionate importance.
\end{itemize}

The geometric mean prevents the gradient term from dominating the score when gradients are noisy or have high variance across layers.

\subsection{Adaptive Layer-Module Allocation}

Following FLAP~\citep{an2024fluctuation}, we adopt an adaptive allocation strategy that distributes pruning budgets across layers and module types based on measured redundancy.

\paragraph{Cross-layer standardization.}
The raw TriSP scores differ in magnitude across layers because deeper layers tend to have different activation scales and gradient norms than shallow ones. Applying a single global threshold to raw scores would prune too aggressively in layers with small scores and too conservatively in layers with large scores. To make scores comparable across layers, we apply per-layer z-score standardization:
\begin{equation}
    \hat{S}_{l,j} = \frac{S_{l,j} - \mu_l}{\sigma_l}
    \label{eq:zscore}
\end{equation}
where $S_{l,j}$ is the TriSP importance score of channel $j$ in layer $l$, and $\mu_l$ and $\sigma_l$ are the mean and standard deviation of all channel scores within layer $l$.

\paragraph{Attention head scoring.}
For multi-head attention, the per-channel TriSP scores of the output projection are grouped by head. Each head spans $d_h$ contiguous channels, and the head-level importance is the mean of its channel scores:
\begin{equation}
    H_{l,k} = \frac{1}{d_h} \sum_{j \in \text{head}_k} \hat{S}_{l,j}
    \label{eq:head_score}
\end{equation}
where $H_{l,k}$ is the importance score of head $k$ in layer $l$, $d_h$ is the head dimension, and $\hat{S}_{l,j}$ are the z-score normalized channel scores from Equation~\ref{eq:zscore}.

\paragraph{Multi-Layer Perceptron neuron scoring.}
Each MLP neuron corresponds to one intermediate dimension, shared across the gate, up, and down projections. The standardized TriSP scores of the down projection's input channels are used directly as per-neuron importance scores.

\paragraph{Global threshold with compression weighting.}
All head scores and neuron scores across all layers are concatenated into a single vector and sorted in descending order. Because removing one attention head eliminates $4 \times d_h \times d_{\text{hidden}}$ parameters (across four projections: $W_Q$, $W_K$, $W_V$, and $W_O$), while removing one MLP neuron eliminates only $3 \times d_{\text{hidden}}$ parameters (across three projections: $W_{\text{gate}}$, $W_{\text{up}}$, and $W_{\text{down}}$), pruning a head removes far more parameters than pruning a neuron. To account for this, each component is assigned a \textit{compression weight} $c_i$ that reflects its relative parameter cost: each attention head receives $c_i = 4 d_h / 3$, while each MLP neuron receives $c_i = 1$.

The global threshold is determined in two steps. First, the optimal cutoff position $k^*$ is found by scanning the sorted list and selecting the position where the cumulative compression weight is closest to the total compression weight that should be retained:
\begin{equation}
    k^* = \underset{k}{\arg\min} \left|
    \sum_{m=1}^{k} c_{\pi(m)} - (1 - p) \sum_i c_i \right|
    \label{eq:kstar}
\end{equation}
where $\pi$ is the permutation that sorts all component scores in descending order, $c_{\pi(m)}$ is the compression weight of the component at sorted position $m$, $p$ is the target pruning ratio, and $(1 - p) \sum_i c_i$ is the total compression weight. Then, the threshold $\tau$ is set to the score at that position:
\begin{equation}
    \tau = S_{\pi(k^*)}
    \label{eq:threshold}
\end{equation}
All components with $S_i > \tau$ are retained; the rest are pruned.

\subsection{Structural Pruning}

Once the binary masks are determined, we compress the model by physically removing the pruned dimensions from the weight matrices, producing a smaller dense model.

The rows of the query projection $W_Q$, key projection $W_K$, and value projection $W_V$ corresponding to pruned heads are removed, and the corresponding columns of the output projection $W_O$ are removed. For the MLP, we remove the rows of the up projection $W_{\text{up}}$ and gate projection $W_{\text{gate}}$ corresponding to pruned neurons, and the corresponding columns of the down projection $W_{\text{down}}$. The number of retained MLP neurons per layer is rounded down to the nearest multiple of 64 to ensure efficient GPU memory access and matrix computation.

In the original FLAP formulation, a fixed bias vector compensates for the removed channels by adding the mean activation of pruned inputs to the layer output: $b_{\text{comp}} = W_{\text{out}}^\top \cdot \bar{x}_{\text{pruned}}$, where $W_{\text{out}}$ is the weight matrix of the output projection and $\bar{x}_{\text{pruned}}$ is the mean activation vector of the pruned channels computed over the calibration set. This is a rank-zero approximation that captures only the mean and ignores per-sample variation.

We deliberately omit this bias term, as the LoRA adapters applied in the recovery stage learn a richer, input-dependent correction that subsumes what the fixed bias provides (see Figure~\ref{fig:recovery}).

\subsection{Low-Rank Adaptation Recovery}

Structural pruning introduces a quality-accuracy trade-off. To recover performance efficiently, we employ Low-Rank Adaptation (LoRA)~\citep{hu2022lora} as a post-pruning recovery step. LoRA injects low-rank adapter matrices into the pruned model's attention projections ($W_Q$, $W_K$, $W_V$, and $W_O$) of every transformer layer. Only the adapter parameters are trained while all pretrained weights remain frozen, making the recovery lightweight. After training, adapters are merged back into the pruned weights.


\begin{algorithm}[t]
\caption{TriSP: Tri-Signal Structured Pruning with LoRA Recovery}
\label{alg:model_pruning}
\footnotesize
\begin{algorithmic}[1]

\Require Model $\mathcal{M}$ with $L$ transformer layers, calibration set $\mathcal{D}_{\text{cal}}$ ($N{=}512$ samples), pruning ratio $p$, LoRA rank $r$, recovery dataset $\mathcal{D}_{\text{rec}}$
\Ensure Pruned and recovered model $\mathcal{M}'$

\State \textbf{Stage 1: Importance Estimation}
\For{each layer $l = 1, \dots, L$}
    \State Initialize $a_{l,j} \gets 0$, \; $\bar{G}_{l,ij} \gets 0$; collect input activations at $W_O^{(l)}$ and $W_{\text{down}}^{(l)}$
\EndFor
\For{each sample $(x, y) \in \mathcal{D}_{\text{cal}}$}
    \State Forward pass: $\hat{y} \gets \mathcal{M}(x)$; compute $\mathcal{L}$ (Eq.~\ref{eq:ce_loss}); backward pass
    \For{each layer $l$}
        \State Accumulate $\bar{G}_{l,ij}$ (Eq.~\ref{eq:grad_avg}); update $a_{l,j}$ (Eq.~\ref{eq:running_norm})
    \EndFor
\EndFor
\For{each layer $l$}
    \State Compute WIFN$_{l,j}$ (Eq.~\ref{eq:wifn}), GS$_{l,j}$ (Eq.~\ref{eq:gs}), TriSP score $S_{l,j}$ (Eq.~\ref{eq:gw})
\EndFor

\State \textbf{Stage 2: Adaptive Layer-Module (AL-AM)}
\For{each layer $l$}
    \State Standardize scores $\hat{S}_{l,j}$ via z-score normalization (Eq.~\ref{eq:zscore})
\EndFor
\State Aggregate per-head importance $H_{l,k}$ (Eq.~\ref{eq:head_score});
\State Sort the scores in descending order
\State Assign compression weights: $c_i \gets 4d_h/3$ for heads, $c_i \gets 1$ for neurons
\State Find global threshold $\tau$ for target ratio $p$ (Eq.~\ref{eq:kstar}); generate masks $M_i \gets \mathbf{1}[\text{score}_i > \tau]$

\State \textbf{Stage 3: Structural Pruning and Recovery}
\For{each layer $l$}
    \State Remove pruned head rows from $W_Q, W_K, W_V$ and columns from $W_O$; remove pruned neuron rows from $W_{\text{up}}, W_{\text{gate}}$ and columns from $W_{\text{down}}$;
\EndFor
\State Freeze all weights; inject LoRA adapters into $W_Q, W_K, W_V, W_O$
\State Fine-tune adapters on $\mathcal{D}_{\text{rec}}$ for 2 epochs; merge adapters into pruned weights
\State \Return $\mathcal{M}'$
\end{algorithmic}
\end{algorithm}

\section{Experiments and Results}
\label{sec:experiments}

\subsection{Experimental Setup}

\paragraph{Models.}
The primary pruning experiments are conducted on LLaMA-2-7B~\citep{touvron2023llama} and Vicuna-7B~\citep{chiang2023vicuna} at pruning ratios of 20\%, 30\%, and 50\%. To assess generalization across architectures, the same pipeline is additionally evaluated on DeepSeek-7B~\citep{bi2024deepseek} and Mistral-7B~\citep{jiang2023mistral}. Table~\ref{tab:model_arch} summarizes the architectural configurations of all models.

\begin{table}[h]
\centering
\footnotesize
\caption{Architectural configurations of all models used in this work.}
\label{tab:model_arch}
\begin{tabular}{lccccccc}
\toprule
\textbf{Model} & $L$ & $d$ & $H_Q$ & $H_{KV}$ & $d_{\text{ff}}$ & $|V|$ & \textbf{Context} \\
\midrule
LLaMA-2-7B   & 32 & 4096 & 32 & 32 & 11\,008 & 32\,000 & 4\,096 \\
Vicuna-7B    & 32 & 4096 & 32 & 32 & 11\,008 & 32\,000 & 4\,096 \\
DeepSeek-7B  & 30 & 4096 & 32 & 32 & 11\,008 & 102\,400 & 4\,096 \\
Mistral-7B   & 32 & 4096 & 32 &  8 & 14\,336 & 32\,000 & 32\,768 \\
\bottomrule
\end{tabular}
\vspace{0.3em}

{\scriptsize $L$: layers, $d$: hidden dim., $H_Q$: query heads, $H_{KV}$: KV heads, $d_{\text{ff}}$: intermediate dim., $|V|$: vocab size.}
\end{table}

\paragraph{Baselines.}
The proposed TriSP pruning pipeline is compared against FLAP~\citep{an2024fluctuation} (with and without bias compensation) and LLM-Pruner~\citep{ma2023llm} (with and without LoRA recovery). All methods use the same calibration set (512 WikiText-2 samples) for fair comparison.

\paragraph{Metrics.}
We evaluate four dimensions: (i)~language modeling quality via WikiText-2 test and Penn Treebank perplexity (PTB); (ii)~downstream task accuracy via zero-shot evaluation on six commonsense benchmarks: PIQA~\citep{bisk2020piqa}, HellaSwag~\citep{zellers2019hellaswag}, WinoGrande~\citep{sakaguchi2021winogrande}, ARC-Easy and ARC-Challenge~\citep{clark2018think}, and OBQA~\citep{mihaylov2018can};
 (iii)~computational cost via MACs; and (iv)~inference efficiency via forward-pass latency and throughput, measured over 100 timed runs at batch size 32 with sequence length 64 on a single NVIDIA H100 80GB GPU.

 \paragraph{LoRA recovery.}
LoRA adapters with rank $r = 8$ and scaling factor $\alpha = 16$ are applied to the four attention projections of every transformer layer. A dropout rate of 0.05 is applied to the LoRA matrices. Recovery fine-tuning is performed on 25{,}000 samples of the Alpaca instruction-tuning dataset~\citep{taori2023alpaca} for 2 epochs, with a learning rate of $10^{-4}$, cosine schedule. Extending LoRA to MLP projections is left for future work, as attention-only recovery was sufficient to achieve the best results at all tested ratios.

\subsection{Language Modeling Perplexity}

\begin{table}[h]
\centering
\footnotesize
\caption{WikiText-2 perplexity of pruning methods across two models and three pruning ratios. Best results in \textbf{bold}.}
\label{tab:ppl}
\begin{tabular}{c l l cc}
\toprule
Ratio & Method & Recovery & LLaMA-7B & Vicuna-7B \\
\midrule
0\% & Dense & -- & 5.47 & 6.78 \\
\midrule
\multirow{4}{*}{20\%}
& \multirow{2}{*}{FLAP}
 & w/o tune  & 7.52 & 9.50 \\
 &  & bias  & 7.11 & 9.04 \\
\cmidrule(lr){2-5}
 & \cellcolor{blue!8} & \cellcolor{blue!8}w/o tune  & \cellcolor{blue!8}7.06 & \cellcolor{blue!8}8.81 \\
 & \multirow{-2}{*}{\cellcolor{blue!8}TriSP} & \cellcolor{blue!8}LoRA & \cellcolor{blue!8}\textbf{6.80} & \cellcolor{blue!8}\textbf{7.77} \\
\midrule
\multirow{4}{*}{30\%}
 & \multirow{2}{*}{FLAP}
 & w/o tune  & 9.87 & 12.72 \\
 &  & bias  & 8.61 & 11.38 \\
\cmidrule(lr){2-5}
 & \cellcolor{blue!8} & \cellcolor{blue!8}w/o tune  & \cellcolor{blue!8}9.03 & \cellcolor{blue!8}10.95 \\
 & \multirow{-2}{*}{\cellcolor{blue!8}TriSP} & \cellcolor{blue!8}LoRA & \cellcolor{blue!8}\textbf{8.05} & \cellcolor{blue!8}\textbf{9.10} \\
\midrule
\multirow{4}{*}{50\%}
 & \multirow{2}{*}{FLAP}
 & w/o tune  & 29.43 & 36.86 \\
 &  & bias  & 16.03 & 22.45 \\
\cmidrule(lr){2-5}
 & \cellcolor{blue!8} & \cellcolor{blue!8}w/o tune  & \cellcolor{blue!8}27.63 & \cellcolor{blue!8}31.67 \\
 & \multirow{-2}{*}{\cellcolor{blue!8}TriSP} & \cellcolor{blue!8}LoRA & \cellcolor{blue!8}\textbf{13.44} & \cellcolor{blue!8}\textbf{14.82} \\
\bottomrule
\end{tabular}
\end{table}

Table~\ref{tab:ppl} reports WikiText-2 perplexity across two models and three pruning ratios. At all ratios and across both models, TriSP with LoRA recovery achieves the lowest perplexity. On LLaMA-7B, TriSP + LoRA reaches 6.80 at 20\%, 8.05 at 30\%, and 13.44 at 50\%, consistently outperforming both FLAP without recovery and FLAP with bias compensation. On Vicuna-7B at 50\%, TriSP + LoRA achieves 14.82 compared to 22.45 for FLAP + bias, a 34\% relative improvement.

\subsection{Zero-Shot Downstream Tasks}


\begin{table}[t]
\centering
\footnotesize
\caption{Zero-shot commonsense reasoning accuracy (\%) on LLaMA-2-7B at 20\% and 50\% pruning. Best results in \textbf{bold}.}
\label{tab:zeroshot}
\resizebox{\textwidth}{!}{%
\begin{tabular}{c l cccccc|c}
\toprule
Ratio & Method & PIQA & HellaSwag & WinoGrande & ARC-e & ARC-c & OBQA & Avg. \\
\midrule
0\% & LLaMA-7B & 78.13 & 74.06 & 69.38 & 75.46 & 43.09 & 42.60 & 63.79 \\
\midrule
\multirow{6}{*}{20\%}
 & LLM-Pruner w/o tune & 74.43 & 61.46 & 59.75 & 68.43 & 34.64 & 39.40 & 56.35 \\
 & LLM-Pruner w/ LoRA & 75.63 & 64.57 & 61.17 & 72.52 & 38.57 & 40.20 & 58.78 \\
 & FLAP w/o bias & 72.80 & 61.51 & 64.01 & 60.31 & 29.61 & 38.80 & 54.51 \\
 & FLAP w/ bias & 73.72 & 63.14 & 64.56 & 64.44 & 32.17 & 40.00 & 56.34 \\
\cmidrule(lr){2-9}
 & \cellcolor{blue!8}TriSP w/o tune & \cellcolor{blue!8}75.90 & \cellcolor{blue!8}67.97 & \cellcolor{blue!8}65.67 & \cellcolor{blue!8}70.54 & \cellcolor{blue!8}36.43 & \cellcolor{blue!8}39.60 & \cellcolor{blue!8}59.35 \\
 & \cellcolor{blue!8}TriSP w/ LoRA & \cellcolor{blue!8}\textbf{76.77} & \cellcolor{blue!8}\textbf{68.20} & \cellcolor{blue!8}\textbf{66.85} & \cellcolor{blue!8}\textbf{73.02} & \cellcolor{blue!8}\textbf{40.02} & \cellcolor{blue!8}\textbf{41.00} & \cellcolor{blue!8}\textbf{60.98} \\
\midrule
\multirow{6}{*}{50\%}
 & LLM-Pruner w/o tune & 53.48 & 27.03 & 48.54 & 30.18 & 20.65 & 24.80 & 34.11 \\
 & LLM-Pruner w/ LoRA & 62.40 & 37.56 & 51.22 & 44.44 & 22.10 & 28.40 & 41.02 \\
 & FLAP w/o bias & 60.01 & 42.56 & 50.75 & 36.78 & 22.95 & 32.00 & 40.84 \\
 & FLAP w/ bias & 60.94 & 40.62 & 53.04 & 43.98 & 22.70 & 33.60 & 42.48 \\
\cmidrule(lr){2-9}
 & \cellcolor{blue!8}TriSP w/o tune & \cellcolor{blue!8}61.48 & \cellcolor{blue!8}42.42 & \cellcolor{blue!8}51.62 & \cellcolor{blue!8}45.92 & \cellcolor{blue!8}25.09 & \cellcolor{blue!8}32.20 & \cellcolor{blue!8}43.12 \\
 & \cellcolor{blue!8}TriSP w/ LoRA & \cellcolor{blue!8}\textbf{67.46} & \cellcolor{blue!8}\textbf{48.17} & \cellcolor{blue!8}\textbf{53.04} & \cellcolor{blue!8}\textbf{56.02} & \cellcolor{blue!8}\textbf{29.78} & \cellcolor{blue!8}\textbf{34.60} & \cellcolor{blue!8}\textbf{48.18} \\
\bottomrule
\end{tabular}%
}
\end{table}

Table~\ref{tab:zeroshot} evaluates zero-shot accuracy on six commonsense reasoning benchmarks for LLaMA-2-7B. At 20\% pruning, TriSP + LoRA achieves the highest average accuracy (60.98\%), outperforming LLM-Pruner + LoRA (58.78\%), FLAP + bias (56.34\%), and FLAP without bias (54.51\%). TriSP + LoRA leads on all six benchmarks, with particularly strong margins on HellaSwag (68.20 vs.\ 64.57 for LLM-Pruner) and WinoGrande (66.85 vs.\ 64.56 for FLAP).

At 50\% pruning, all methods degrade substantially, but TriSP + LoRA retains the best average performance (48.18\%), with notable margins over all baselines, including LLM-Pruner + LoRA on PIQA (+5.06), ARC-e (+11.58), and HellaSwag (+10.61).

\FloatBarrier
\subsection{Generalization to Other Architectures}

To validate that TriSP generalizes beyond the LLaMA-2 family, we evaluate on DeepSeek-7B and Mistral-7B, which differ in several architectural dimensions detailed in Table~\ref{tab:model_arch}.


\begin{table}[t]
\centering
\footnotesize
\caption{Perplexity and zero-shot accuracy (\%) of pruning methods for DeepSeek-7B.}
\label{tab:deepseek}
\resizebox{\textwidth}{!}{%

\begin{tabular}{c l cc|cccccc|c}
\toprule
Ratio & Method & WikiText-2 $\downarrow$ & PTB $\downarrow$ & PIQA & HellaSwag & WinoGrande & ARC-e & ARC-c & OBQA & Avg. $\uparrow$ \\
\midrule
0\% & Dense & 6.84 & 9.27 & 78.51 & 73.85 & 70.48 & 74.58 & 42.32 & 45.20 & 64.16 \\
\midrule
\multirow{4}{*}{20\%}
 & LLM-Pruner w/ LoRA & 9.79 & 14.96 & 77.20 & 66.12 & 63.14 & 70.41 & 36.26 & 41.00 & 59.02 \\
 & FLAP w/ bias & 8.30 & 11.58 & 76.39 & 66.58 & \textbf{65.51} & 67.89 & 35.84 & 43.80 & 59.34 \\
\cmidrule(lr){2-11}
 & \cellcolor{blue!8}TriSP w/o tune & \cellcolor{blue!8}8.53 & \cellcolor{blue!8}12.19 & \cellcolor{blue!8}77.31 & \cellcolor{blue!8}\textbf{70.32} & \cellcolor{blue!8}64.40 & \cellcolor{blue!8}68.98 & \cellcolor{blue!8}37.63 & \cellcolor{blue!8}43.60 & \cellcolor{blue!8}60.37 \\
 & \cellcolor{blue!8}TriSP w/ LoRA & \cellcolor{blue!8}\textbf{8.27} & \cellcolor{blue!8}\textbf{11.92} & \cellcolor{blue!8}\textbf{77.80} & \cellcolor{blue!8}69.87 & \cellcolor{blue!8}65.04 & \cellcolor{blue!8}\textbf{72.60} & \cellcolor{blue!8}\textbf{41.81} & \cellcolor{blue!8}\textbf{44.60} & \cellcolor{blue!8}\textbf{61.95} \\
\midrule
\multirow{4}{*}{30\%}
 & LLM-Pruner w/ LoRA & 15.89 & 38.27 & 74.32 & 58.10 & 56.99 & 64.02 & 31.74 & 38.40 & 53.93 \\
 & FLAP w/ bias & 10.00 & \textbf{13.98} & 73.39 & 59.85 & \textbf{63.06} & 62.04 & 33.45 & 40.40 & 55.37 \\
\cmidrule(lr){2-11}
 & \cellcolor{blue!8}TriSP w/o tune & \cellcolor{blue!8}10.68 & \cellcolor{blue!8}15.54 & \cellcolor{blue!8}75.19 & \cellcolor{blue!8}\textbf{65.58} & \cellcolor{blue!8}60.54 & \cellcolor{blue!8}64.86 & \cellcolor{blue!8}34.98 & \cellcolor{blue!8}42.60 & \cellcolor{blue!8}57.29 \\
 & \cellcolor{blue!8}TriSP w/ LoRA & \cellcolor{blue!8}\textbf{9.82} & \cellcolor{blue!8}14.34 & \cellcolor{blue!8}\textbf{75.68} & \cellcolor{blue!8}65.43 & \cellcolor{blue!8}62.04 & \cellcolor{blue!8}\textbf{68.86} & \cellcolor{blue!8}\textbf{36.52} & \cellcolor{blue!8}\textbf{43.20} & \cellcolor{blue!8}\textbf{58.62} \\
\bottomrule
\end{tabular}%

}
\end{table}

\begin{table}[t]
\centering
\footnotesize
\caption{Perplexity and zero-shot accuracy (\%) of pruning methods for Mistral-7B.}
\label{tab:mistral}
\resizebox{\textwidth}{!}{%
\begin{tabular}{c l cc|cccccc|c}
\toprule
Ratio & Method & WikiText-2 $\downarrow$ & PTB $\downarrow$ & PIQA & HellaSwag & WinoGrande & ARC-e & ARC-c & OBQA & Avg. $\uparrow$ \\
\midrule
0\% & Dense & 5.25 & 26.96 & 80.25 & 78.75 & 74.66 & 80.22 & 49.83 & 46.60 & 68.39 \\
\midrule
\multirow{4}{*}{20\%}
 & LLM-Pruner w/ LoRA & 8.14 & 30.81 & 75.68 & 66.97 & 63.38 & 71.17 & 41.47 & 41.40 & 60.01 \\
 & FLAP w/ bias & 6.46 & 42.14 & 69.10 & 52.22 & 66.54 & 66.71 & 33.79 & 40.80 & 54.86 \\
\cmidrule(lr){2-11}
 & \cellcolor{blue!8}TriSP w/o tune & \cellcolor{blue!8}\textbf{6.15} & \cellcolor{blue!8}34.39 & \cellcolor{blue!8}74.32 & \cellcolor{blue!8}61.94 & \cellcolor{blue!8}\textbf{69.06} & \cellcolor{blue!8}70.83 & \cellcolor{blue!8}39.51 & \cellcolor{blue!8}41.40 & \cellcolor{blue!8}59.51 \\
 & \cellcolor{blue!8}TriSP w/ LoRA & \cellcolor{blue!8}6.35 & \cellcolor{blue!8}\textbf{32.83} & \cellcolor{blue!8}\textbf{76.88} & \cellcolor{blue!8}\textbf{67.49} & \cellcolor{blue!8}67.56 & \cellcolor{blue!8}\textbf{74.87} & \cellcolor{blue!8}\textbf{45.82} & \cellcolor{blue!8}\textbf{43.20} & \cellcolor{blue!8}\textbf{62.64} \\
\midrule
\multirow{4}{*}{30\%}
 & LLM-Pruner w/ LoRA & 11.72 & 52.62 & \textbf{72.58} & 59.20 & 58.09 & 63.51 & 33.87 & 38.20 & 54.24 \\
 & FLAP w/ bias & 7.78 & 57.22 & 65.56 & 44.78 & 61.64 & 59.85 & 28.84 & 37.60 & 49.71 \\
\cmidrule(lr){2-11}
 & \cellcolor{blue!8}TriSP w/o tune & \cellcolor{blue!8}7.26 & \cellcolor{blue!8}43.31 & \cellcolor{blue!8}67.79 & \cellcolor{blue!8}51.97 & \cellcolor{blue!8}64.25 & \cellcolor{blue!8}65.78 & \cellcolor{blue!8}32.94 & \cellcolor{blue!8}39.20 & \cellcolor{blue!8}53.66 \\
 & \cellcolor{blue!8}TriSP w/ LoRA & \cellcolor{blue!8}\textbf{7.04} & \cellcolor{blue!8}\textbf{38.27} & \cellcolor{blue!8}72.47 & \cellcolor{blue!8}\textbf{61.29} & \cellcolor{blue!8}\textbf{65.67} & \cellcolor{blue!8}\textbf{69.53} & \cellcolor{blue!8}\textbf{38.14} & \cellcolor{blue!8}\textbf{41.00} & \cellcolor{blue!8}\textbf{58.02} \\
\bottomrule
\end{tabular}%
}
\end{table}

\paragraph{DeepSeek-7B.}
Table~\ref{tab:deepseek} reports perplexity and zero-shot accuracy at 20\% and 30\% pruning. TriSP + LoRA achieves the best WikiText-2 perplexity at both ratios (8.27 at 20\%, 9.82 at 30\%) and the highest average zero-shot accuracy (61.95\% at 20\%, 58.62\% at 30\%). At 20\%, TriSP + LoRA improves over FLAP + bias by 2.62 points in average accuracy and over LLM-Pruner + LoRA by 2.93 points. The gap widens at 30\%, where TriSP leads FLAP + bias by 3.26 points.

\paragraph{Mistral-7B.}
Table~\ref{tab:mistral} shows results on Mistral-7B. TriSP + LoRA achieves the best average accuracy at both ratios (62.64\% at 20\%, 58.02\% at 30\%). At 30\%, TriSP outperforms FLAP + bias by 8.31 points (58.02 vs.\ 49.71) and LLM-Pruner + LoRA by 3.78 points. Notably, TriSP without LoRA achieves the lowest WikiText-2 perplexity at 20\% (6.15 vs.\ 6.35 for TriSP + LoRA), however, LoRA still improves average downstream accuracy (62.64 vs.\ 59.51).

\subsection{Inference Efficiency}

\begin{table}[t]
\centering
\footnotesize
\caption{Inference statistics of compressed LLaMA-2-7B. Measured with sequence length 64, batch size 32 on a single NVIDIA H100.}
\label{tab:inference_stats}
\begin{tabular}{lcccccc}
\toprule
\textbf{Method} & \textbf{Ratio} & \textbf{Params} & \textbf{MACs} & \textbf{Memory} & \textbf{Latency (ms)} & \textbf{Tokens/s} \\
\midrule
LLaMA-2-7B          & 0\%  & 6.74B & 422.85G & 12885 MiB & 49.83              & 41.1k \\
\midrule
LLM-Pruner           & 20\% & 5.47B & 341.67G & 10433 MiB & 50.47 ($\approx$0\%) & 40.6k \\
FLAP                 & 20\% & 5.44B & 340.06G & 10424 MiB & 49.40 ($\approx$0\%) & 41.5k \\
\rowcolor{blue!8} \textbf{TriSP (Ours)}   & 20\% & 5.43B & 339.35G & 10462 MiB & 40.94 ($\downarrow$18\%) & 50.0k \\
\midrule
LLM-Pruner           & 50\% & 3.50B & 215.62G & 6677 MiB & \underline{26.11} ($\downarrow$48\%) & 78.4k \\
FLAP                 & 50\% & 3.50B & 215.67G & 6726 MiB  & 34.02 ($\downarrow$32\%) & 60.2k \\
\rowcolor{blue!8} \textbf{TriSP (Ours)}   & 50\% & 3.49B & 214.82G & 6751 MiB  & 27.46 ($\downarrow$45\%) & 74.6k \\
\bottomrule
\end{tabular}
\end{table}

Table~\ref{tab:inference_stats} reports inference statistics for LLaMA-2-7B models. At 20\% pruning, all three methods achieve comparable parameter counts ($\sim$5.4B) and MACs ($\sim$340G). The key differentiator is latency: TriSP achieves an 18\% latency reduction (40.94\,ms vs.\ 49.83\,ms for dense) and a 22\% throughput improvement over the dense baseline (50.0k\,tokens/s vs.\ 41.1k), while both LLM-Pruner (50.47\,ms) and FLAP (49.40\,ms) show weak speedup. This difference is likely because TriSP rounds MLP dimensions to multiples of 64, ensuring efficient GPU memory access and matrix computation, while LLM-Pruner and FLAP produce dimensions that do not align with the hardware execution granularity.

At 50\%, LLM-Pruner achieves the lowest latency (26.11\,ms, $\downarrow$48\%), with TriSP close behind (27.46\,ms, $\downarrow$45\%), while FLAP lags at 34.02\,ms ($\downarrow$32\%). TriSP delivers 74.6k tokens/s, an 82\% throughput improvement over the dense baseline, compared to 78.4k ($\uparrow$91\%) for LLM-Pruner and 60.2k ($\uparrow$46\%) for FLAP. Memory savings scale proportionally with parameter reduction, dropping from 12{,}885\,MiB to $\sim$6{,}700\,MiB at 50\%.

\subsection{Ablation Studies}

We systematically examine three components of the TriSP pipeline: the recovery strategy, the importance metric, and the allocation strategy.

\subsubsection{Effect of Recovery Strategy}

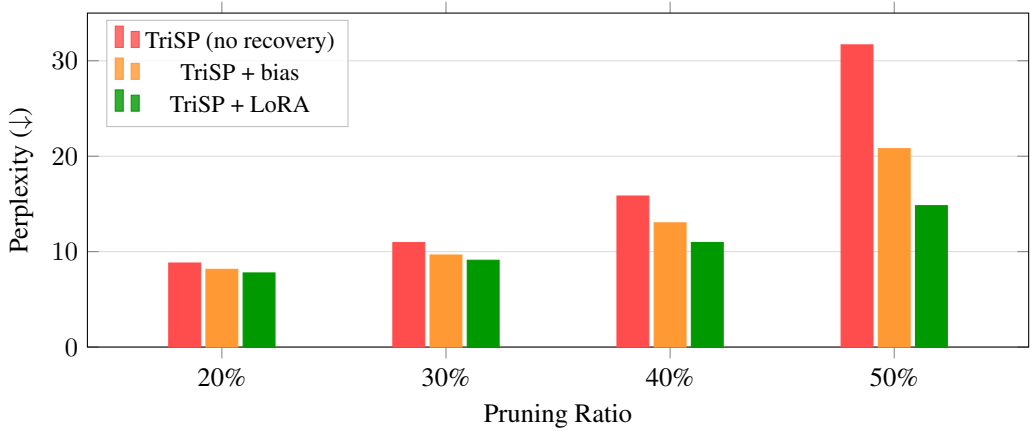
\begin{figure}[t]
\centering
\begin{tikzpicture}
\begin{axis}[
    ybar,
    width=0.85\textwidth,
    height=6cm,
    bar width=12pt,
    ylabel={Perplexity ($\downarrow$)},
    xlabel={Pruning Ratio},
    symbolic x coords={20\%,30\%,40\%,50\%},
    xtick=data,
    enlarge x limits=0.2,
    ymin=0,
    ymax=35,
    legend style={
        at={(0.02,0.98)},
        anchor=north west,
        legend columns=1,
        font=\small,
        draw=gray!50,
        fill=white,
        fill opacity=0.8,
        text opacity=1,
    },
    grid=major,
    ymajorgrids=true,
    xmajorgrids=false,
    grid style={line width=0.3pt, draw=gray!30},
]

\addplot[fill=red!70, draw=red!70]
    coordinates {(20\%,8.81) (30\%,10.95) (40\%,15.83) (50\%,31.67)};

\addplot[fill=orange!80, draw=orange!80]
    coordinates {(20\%,8.15) (30\%,9.65) (40\%,13.02) (50\%,20.79)};

\addplot[fill=green!60!black, draw=green!60!black]
    coordinates {(20\%,7.77) (30\%,9.10) (40\%,10.96) (50\%,14.82)};

\legend{TriSP (no recovery), TriSP + bias, TriSP + LoRA}

\end{axis}
\end{tikzpicture}
\caption{Vicuna-7B: WikiText-2 perplexity under TriSP pruning with different recovery strategies.}
\label{fig:recovery}
\end{figure}

Figure~\ref{fig:recovery} compares three recovery strategies for TriSP pruning on Vicuna-7B: no recovery, bias compensation, and LoRA fine-tuning. LoRA recovery performs best across all pruning ratios, reducing perplexity from 8.81 to 7.77 at 20\%, from 10.95 to 9.10 at 30\%, from 15.83 to 10.96 at 40\%, and from 31.67 to 14.82 at 50\%. Bias compensation provides intermediate improvements at moderate ratios (e.g., 9.65 at 30\% and 13.02 at 40\%), but its advantage over no recovery diminishes relative to LoRA as compression increases. At 50\%, the gap between LoRA and bias widens sharply (14.82 vs.\ 20.79). At 40\%, LoRA reduces perplexity by 30.7\% relative to no recovery, while bias compensation achieves only 17.8\%. This confirms that LoRA's input-dependent, rank-$r$ correction subsumes and exceeds what a rank-zero fixed bias term can provide.

\subsubsection{Importance Metric Ablation}

Table~\ref{tab:metric-ablation} isolates the contribution of each component in the TriSP metric by evaluating WIFN alone, GS alone, and the combined TriSP score on Vicuna-7B. At all pruning ratios, the combined TriSP metric outperforms either component in isolation: at $p=0.4$, TriSP achieves 15.83 compared to 16.12 for WIFN and 17.14 for GS. The gap for the GS component grows from 0.10 at $p=0.2$ to 1.31 at $p=0.4$, suggesting that gradient sensitivity alone overprunes certain channels that WIFN's activation-weighted magnitude signal protects.

\begin{table}[t]
\centering
\caption{Metric ablation on Vicuna-7B: WikiText-2 perplexity using individual importance signals vs.\ the full TriSP metric.}
\label{tab:metric-ablation}
\begin{tabular}{l ccc}
\toprule
Metric & $p=0.2$ & $p=0.3$ & $p=0.4$ \\
\midrule
WIFN only & 8.92 & 11.26 & 16.12 \\
GS only & 8.91 & 12.17 & 17.14 \\
\rowcolor{blue!8} TriSP (ours) & \textbf{8.81} & \textbf{10.95} & \textbf{15.83} \\
\bottomrule
\end{tabular}
\end{table}

\subsubsection{Allocation Strategy Ablation}

\begin{figure}[t]
\centering
\begin{tikzpicture}
\begin{axis}[
    ybar,
    width=0.85\textwidth,
    height=6cm,
    bar width=10pt,
    ylabel={WikiText-2 Perplexity ($\downarrow$)},
    symbolic x coords={0.2, 0.3, 0.4, 0.5},
    xtick=data,
    xlabel={Pruning Ratio},
    ymin=0, ymax=60,
    legend style={
        at={(0.02,0.98)},
        anchor=north west,
        legend columns=1,
        font=\small,
        draw=gray!50,
        fill=white,
        fill opacity=0.8,
        text opacity=1,
    },
    grid=major,
    ymajorgrids=true,
    xmajorgrids=false,
    grid style={line width=0.2pt, draw=gray!20},
]

\addplot[fill=blue!50, draw=blue!70]
    coordinates {(0.2,7.40) (0.3,9.18) (0.4,22.11) (0.5,53.46)};

\addplot[fill=red!60, draw=red!80]
    coordinates {(0.2,7.06) (0.3,9.03) (0.4,13.84) (0.5,27.63)};

\addplot[fill=green!50, draw=green!70]
    coordinates {(0.2,0.34) (0.3,0.15) (0.4,8.27) (0.5,25.83)};

\legend{UL-UM (uniform), AL-AM (adaptive), $\Delta$ (Gap)}

\end{axis}
\end{tikzpicture}
\caption{WikiText-2 perplexity: uniform (UL-UM) vs.\ adaptive (AL-AM) allocation on LLaMA-2-7B.}
\label{fig:allocation-ablation}
\end{figure}
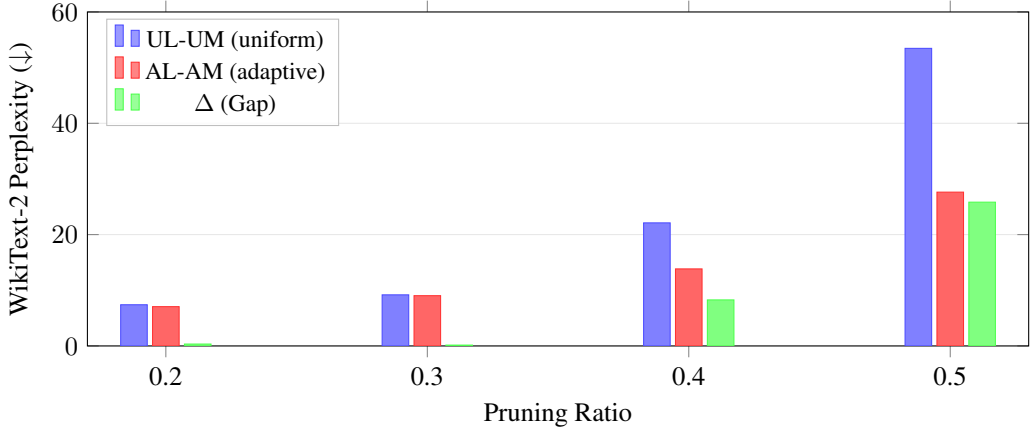

Figure~\ref{fig:allocation-ablation} compares uniform layer-module allocation (UL-UM), which assigns the same pruning ratio to every layer and module type, against adaptive layer-module allocation (AL-AM) on LLaMA-2-7B. At mild pruning ($p \leq 0.3$), the gap is negligible ($\Delta \leq 0.34$ perplexity points). Beyond 30\%, the benefit grows sharply: at $p=0.4$, AL-AM reduces perplexity by 8.27 points (13.84 vs.\ 22.11), and at $p=0.5$, the gap reaches 25.83 points (27.63 vs.\ 53.46). This demonstrates that adaptive budget redistribution becomes critical at aggressive pruning, where uniform allocation causes catastrophic over-pruning in sensitive layers.

\subsection{Discussion}

\paragraph{Why TriSP outperforms FLAP.}
FLAP uses activation variance as a statistical proxy for importance; TriSP includes $|W_{ij}| \cdot \bar{G}_{ij}$ from the Taylor expansion, which directly measures first-order loss sensitivity. This explains why TriSP without recovery already outperforms FLAP without recovery at all ratios: the metric itself makes better pruning decisions, not just the recovery strategy.

\paragraph{Why TriSP outperforms LLM-Pruner.} LLM-Pruner models structural dependencies through dependency graphs, while TriSP treats channels independently. Despite this, TriSP without recovery achieves higher average zero-shot accuracy than LLM-Pruner without recovery (59.35 vs.\ 56.35 at 20\%) and even surpasses LLM-Pruner with LoRA (58.78), indicating that the TriSP importance criterion itself makes better pruning decisions before any recovery is applied. LLM-Pruner's dependency graph constrains which structures can be removed together, which may force the removal of important channels that happen to be coupled with unimportant ones.

\paragraph{Recovery adequacy.}
At mild pruning (20\%), bias compensation and LoRA yield similar improvements, because few channels are removed and the rank-zero mean approximation is sufficient. At aggressive pruning (40--50\%), LoRA's rank-$r$ input-dependent correction vastly outperforms the fixed bias, because a fixed offset (bias) cannot capture the input-dependent variation lost when many channels are removed.

\paragraph{Practical deployment recommendations.}
At 20--30\% pruning, perplexity increases by 1--3 points and zero-shot accuracy drops by 2--6 points, with 14--24\% latency reduction. This operating range is suitable for most applications where modest quality loss is acceptable for cost reduction. At 50\% pruning, perplexity roughly doubles and zero-shot accuracy drops by $\sim$15 points, with 42--45\% latency reduction. This aggressive regime is appropriate only for latency-critical applications with high quality tolerance.

\section{Conclusion}
\label{sec:conclusion}

We presented TriSP (Tri-Signal Structured Pruning), a method that combines activation-weighted magnitude with gradient sensitivity via a geometric mean to produce channel-level importance scores. Paired with adaptive layer-module (AL-AM) allocation and LoRA recovery, TriSP achieves the lowest perplexity and highest downstream accuracy across four models (LLaMA-2-7B, Vicuna-7B, DeepSeek-7B, Mistral-7B) at all tested pruning ratios, while delivering 22--82\% throughput improvements through hardware-aligned pruning.

The three ablation studies confirm that each component contributes independently: the TriSP metric outperforms either WIFN or GS alone, AL-AM allocation becomes critical at aggressive pruning (reducing perplexity by 25.83 points at 50\%), and LoRA recovery outperforms bias compensation by an increasing margin as pruning intensifies (30.7\% vs.\ 17.8\% perplexity reduction at 40\%). The systematic comparison between bias compensation and LoRA recovery confirms that input-dependent correction is essential at aggressive compression ratios.

Current limitations include the restriction to decoder-only transformer architectures at 7B scale, and evaluation limited to English-language benchmarks. Future work includes extending LoRA to MLP projections for better recovery, scaling to larger models (30B+), evaluating on broader multilingual benchmarks, and applying TriSP to vision-language models (VLMs).


\bibliographystyle{plainnat}
\bibliography{references}

\appendix

\section{Perplexity Scaling}
\label{app:ppl-scaling}

\begin{figure}[h]
\centering
\begin{tikzpicture}
\begin{axis}[
    width=0.85\textwidth,
    height=7cm,
    xlabel={Pruning Ratio (\%)},
    ylabel={WikiText-2 Perplexity ($\downarrow$)},
    xtick={0,20,30,50},
    xmin=-2, xmax=55,
    ymin=4, ymax=32,
    legend style={
        at={(0.02,0.98)},
        anchor=north west,
        font=\small,
        draw=gray!50,
    },
    grid=major,
    grid style={line width=0.2pt, draw=gray!20},
    mark size=3pt,
    thick,
]

\addplot[color=blue!60, mark=triangle, dashed, mark options={solid}]
    coordinates {(0,5.47) (20,7.52) (30,9.87) (50,29.43)};

\addplot[color=blue!90, mark=triangle*, mark options={solid}]
    coordinates {(0,5.47) (20,7.11) (30,8.61) (50,16.03)};

\addplot[color=red!60, mark=square, dashed, mark options={solid}]
    coordinates {(0,5.47) (20,7.06) (30,9.03) (50,27.63)};

\addplot[color=red!90, mark=square*, mark options={solid}]
    coordinates {(0,5.47) (20,6.80) (30,8.05) (50,13.44)};

\legend{FLAP w/o bias, FLAP w/ bias, TriSP w/o tune, TriSP + LoRA}

\end{axis}
\end{tikzpicture}
\caption{WikiText-2 perplexity scaling with pruning ratio on
LLaMA-2-7B. TriSP + LoRA maintains the lowest perplexity at all
ratios, with the gap widening significantly at 50\% pruning
(13.44 vs.\ 16.03 for FLAP + bias).}
\label{fig:ppl-scaling}
\end{figure}
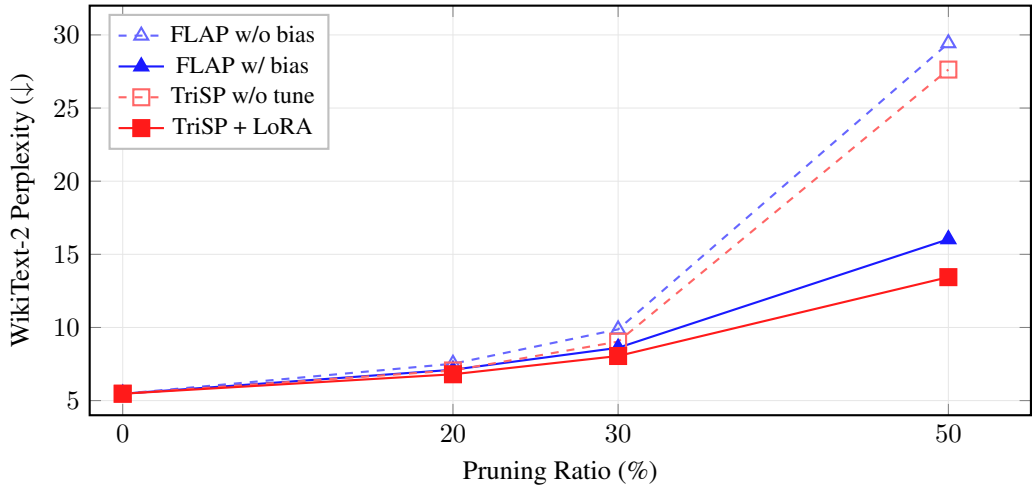

Figure~\ref{fig:ppl-scaling} visualizes the perplexity
scaling behavior on LLaMA-2-7B across pruning ratios.
Both FLAP and TriSP degrade gracefully at moderate ratios,
but diverge sharply at 50\%: TriSP + LoRA maintains 13.44
while FLAP + bias rises to 16.03. TriSP without any recovery
(27.63) already outperforms FLAP without recovery (29.43).

\section{Inference Throughput}
\label{app:inference}

\begin{figure}[t]
\centering
\begin{tikzpicture}
\begin{axis}[
    ybar,
    width=0.85\textwidth,
    height=6.5cm,
    bar width=14pt,
    ylabel={Throughput (kTokens/s)},
    symbolic x coords={Dense, {LLM-P 20\%}, {FLAP 20\%}, {TriSP 20\%}, {LLM-P 50\%}, {FLAP 50\%}, {TriSP 50\%}},
    xtick=data,
    x tick label style={font=\scriptsize, rotate=30, anchor=east},
    enlarge x limits=0.08,
    ymin=0, ymax=95,
    nodes near coords,
    every node near coord/.style={font=\scriptsize, above, yshift=1pt},
    grid=major,
    ymajorgrids=true,
    xmajorgrids=false,
    grid style={line width=0.2pt, draw=gray!20},
    extra y ticks={41.1},
    extra y tick labels={},
    extra y tick style={grid style={line width=0.8pt, draw=gray!50, dashed}},
]

\addplot[fill=gray!40, draw=red!60]
    coordinates {(Dense,41.1) ({LLM-P 20\%},40.6) ({FLAP 20\%},41.5) ({TriSP 20\%},50.0) ({LLM-P 50\%},78.4) ({FLAP 50\%},60.2) ({TriSP 50\%},74.6)};

\end{axis}
\end{tikzpicture}
\caption{Inference throughput of LLaMA-2-7B (seq=64, batch=32,
H100). At 20\% pruning, TriSP achieves 50.0k tokens/s
($\uparrow$22\%) while LLM-Pruner and FLAP show no meaningful
speedup. At 50\%, TriSP delivers 74.6k tokens/s ($\uparrow$82\%)
with 45\% latency reduction.}
\label{fig:inference}
\end{figure}
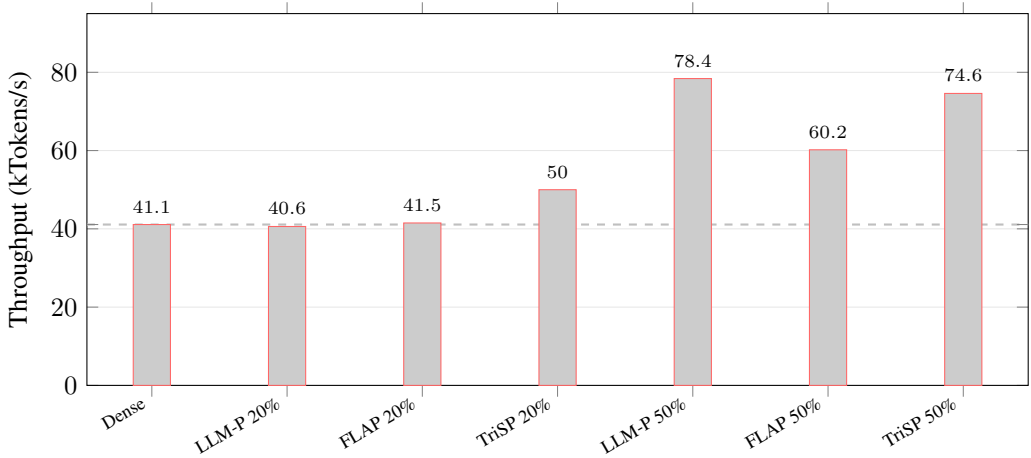

Figure~\ref{fig:inference} reports inference throughput for
compressed LLaMA-2-7B models. At 20\% pruning, TriSP achieves
50.0k tokens/s ($\uparrow$22\%), while LLM-Pruner and FLAP
show no meaningful speedup. At 50\%, TriSP delivers 74.6k
tokens/s with 45\% latency reduction.

\section{DeepSeek-7B Inference Analysis}
\label{app:deepseek-inference}

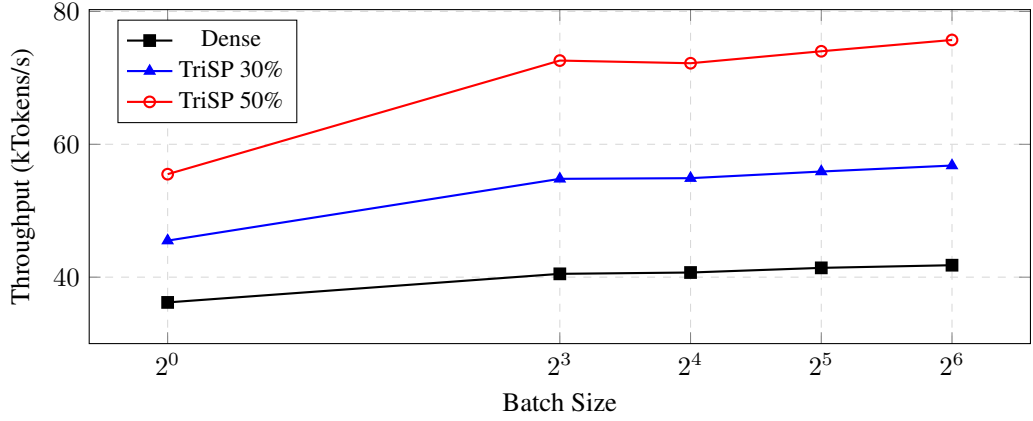
\begin{figure}[t]
\centering
\begin{tikzpicture}
\begin{axis}[
    width=0.85\textwidth,
    height=6cm,
    xlabel={Batch Size},
    ylabel={Throughput (kTokens/s)},
    xtick={1,8,16,32,64},
    xmode=log,
    log basis x=2,
    ymin=30,
    legend style={at={(0.03,0.97)}, anchor=north west, font=\small},
    grid=major,
    grid style={dashed, gray!30},
]

\addplot[
    color=black,
    mark=square*,
    thick,
] coordinates {
    (1, 36.2)
    (8, 40.5)
    (16, 40.7)
    (32, 41.4)
    (64, 41.8)
};
\addlegendentry{Dense}

\addplot[
    color=blue,
    mark=triangle*,
    thick,
] coordinates {
    (1, 45.5)
    (8, 54.8)
    (16, 54.9)
    (32, 55.9)
    (64, 56.8)
};
\addlegendentry{TriSP 30\%}

\addplot[
    color=red,
    mark=o,
    thick,
] coordinates {
    (1, 55.5)
    (8, 72.6)
    (16, 72.2)
    (32, 74.0)
    (64, 75.7)
};
\addlegendentry{TriSP 50\%}

\end{axis}
\end{tikzpicture}
\caption{Throughput vs.\ batch size for TriSP-pruned DeepSeek-7B
at sequence length 512. Structured pruning yields consistent
throughput gains that scale with batch size in the
compute-bound regime.}
\label{fig:deepseek_throughput_batch}
\end{figure}

Table~\ref{tab:inference_stats_deepseek} extends the inference
analysis to DeepSeek-7B. TriSP delivers the lowest latency at
20\% and 30\%. Figure~\ref{fig:deepseek_throughput_batch} shows
that the throughput gap widens with batch size, and
Table~\ref{tab:deepseek_speedup} quantifies the speedup at
sequence length 512.

\begin{table}[t]
\centering
\caption{Statistics of the compressed DeepSeek-7B model.
Inference is conducted with a sequence of 64 tokens at batch
size 32 on a single NVIDIA H100.}
\label{tab:inference_stats_deepseek}
\resizebox{\textwidth}{!}{%
\begin{tabular}{lcccccc}
\toprule
\textbf{Method} & \textbf{Ratio} & \textbf{Params} & \textbf{MACs} & \textbf{Memory} & \textbf{Latency (ms)} & \textbf{Tokens/s} \\
\midrule
DeepSeek-7B          & 0\%  & 6.91B & 415.40G & 13181 MiB & 47.90 & 42.8k \\
\midrule
LLM-Pruner           & 20\% & 5.72B & 339.29G & 10912 MiB & 46.26 ($\downarrow$3\%) & 44.3k \\
FLAP                 & 20\% & 5.70B & 337.75G & 10914 MiB & 45.13 ($\downarrow$6\%) & 45.4k \\
\rowcolor{blue!8} \textbf{TriSP (Ours)} & 20\% & 5.69B & 337.04G & 10917 MiB & 41.00 ($\downarrow$14\%) & 50.0k \\
\midrule
LLM-Pruner           & 30\% & 5.13B & 301.24G & 9778 MiB  & 43.30 ($\downarrow$10\%) & 47.3k \\
FLAP                 & 30\% & 5.09B & 298.93G & 9753 MiB  & 44.15 ($\downarrow$8\%) & 46.4k \\
\rowcolor{blue!8} \textbf{TriSP (Ours)} & 30\% & 5.08B & 298.16G & 9797 MiB  & 36.47 ($\downarrow$24\%) & 56.2k \\
\midrule
LLM-Pruner           & 50\% & 3.87B & 221.12G & 7391 MiB  & \underline{26.45} ($\downarrow$45\%) & 77.4k \\
FLAP                 & 50\% & 3.88B & 221.20G & 7442 MiB  & 35.30 ($\downarrow$26\%) & 58.0k \\
\rowcolor{blue!8} \textbf{TriSP (Ours)} & 50\% & 3.87B & 220.50G & 7470 MiB  & 27.97 ($\downarrow$42\%) & 73.2k \\
\bottomrule
\end{tabular}%
}
\end{table}

\begin{table}[t]
\centering
\caption{Speedup of TriSP-pruned DeepSeek-7B relative to the
dense baseline at sequence length 512.}
\label{tab:deepseek_speedup}
\begin{tabular}{c c c}
\toprule
Batch Size & TriSP 30\% & TriSP 50\% \\
\midrule
1  & 1.26$\times$ & 1.53$\times$ \\
8  & 1.35$\times$ & 1.79$\times$ \\
16 & 1.35$\times$ & 1.77$\times$ \\
32 & 1.35$\times$ & 1.79$\times$ \\
64 & 1.36$\times$ & 1.81$\times$ \\
\bottomrule
\end{tabular}
\end{table}

\section{Per-Layer Retention under AL-AM}
\label{app:layer-sparsity}

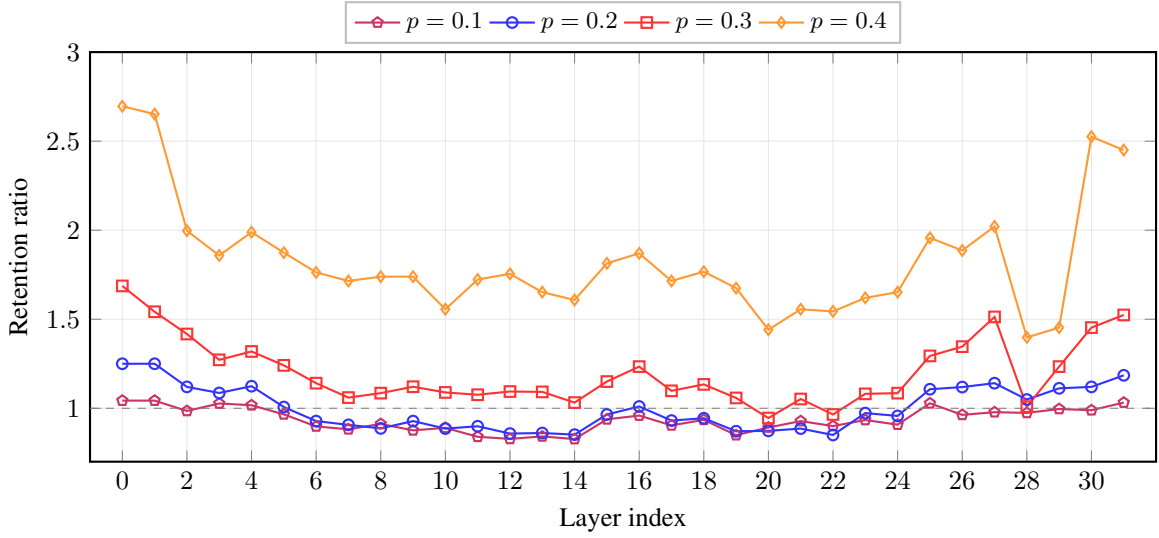
\begin{figure}[t]
\centering
\begin{tikzpicture}
\begin{axis}[
    width=0.95\textwidth,
    height=7cm,
    xlabel={Layer index},
    ylabel={Retention ratio},
    xmin=-1, xmax=32,
    ymin=0.7, ymax=3.0,
    xtick={0,2,4,6,8,10,12,14,16,18,20,22,24,26,28,30},
    legend style={
        at={(0.5,1.02)},
        anchor=south,
        legend columns=5,
        font=\small,
        draw=gray!50,
    },
    grid=major,
    grid style={line width=0.2pt, draw=gray!20},
    mark size=2pt,
    thick,
]

\addplot[color=purple!80, mark=pentagon, mark options={solid, fill=purple!30}]
    coordinates {
        (0,1.043) (1,1.043) (2,0.985) (3,1.027) (4,1.017) (5,0.966)
        (6,0.898) (7,0.882) (8,0.912) (9,0.876) (10,0.890) (11,0.840)
        (12,0.828) (13,0.842) (14,0.827) (15,0.938) (16,0.959) (17,0.904)
        (18,0.935) (19,0.849) (20,0.892) (21,0.928) (22,0.900) (23,0.934)
        (24,0.908) (25,1.028) (26,0.963) (27,0.978) (28,0.974) (29,0.996)
        (30,0.989) (31,1.032)
    };

\addplot[color=blue!80, mark=o, mark options={solid, fill=blue!30}]
    coordinates {
        (0,1.250) (1,1.250) (2,1.120) (3,1.086) (4,1.124) (5,1.007)
        (6,0.928) (7,0.907) (8,0.887) (9,0.928) (10,0.886) (11,0.899)
        (12,0.858) (13,0.861) (14,0.852) (15,0.965) (16,1.010) (17,0.931)
        (18,0.944) (19,0.871) (20,0.873) (21,0.886) (22,0.850) (23,0.973)
        (24,0.957) (25,1.107) (26,1.119) (27,1.141) (28,1.049) (29,1.112)
        (30,1.120) (31,1.185)
    };

\addplot[color=red!80, mark=square, mark options={solid, fill=red!30}]
    coordinates {
        (0,1.687) (1,1.542) (2,1.417) (3,1.272) (4,1.319) (5,1.241)
        (6,1.141) (7,1.060) (8,1.085) (9,1.121) (10,1.089) (11,1.076)
        (12,1.094) (13,1.092) (14,1.032) (15,1.150) (16,1.234) (17,1.098)
        (18,1.134) (19,1.058) (20,0.945) (21,1.052) (22,0.965) (23,1.081)
        (24,1.085) (25,1.294) (26,1.346) (27,1.513) (28,1.009) (29,1.234)
        (30,1.453) (31,1.524)
    };

\addplot[color=orange!80, mark=diamond, mark options={solid, fill=orange!30}]
    coordinates {
        (0,2.696) (1,2.652) (2,1.997) (3,1.858) (4,1.989) (5,1.874)
        (6,1.763) (7,1.715) (8,1.739) (9,1.739) (10,1.556) (11,1.723)
        (12,1.755) (13,1.652) (14,1.608) (15,1.814) (16,1.870) (17,1.715)
        (18,1.767) (19,1.675) (20,1.441) (21,1.556) (22,1.544) (23,1.620)
        (24,1.652) (25,1.957) (26,1.886) (27,2.021) (28,1.398) (29,1.453)
        (30,2.525) (31,2.450)
    };

\addplot[color=gray, dashed, thin, forget plot]
    coordinates {(-1,1.0) (32,1.0)};

\legend{$p = 0.1$, $p = 0.2$, $p = 0.3$, $p = 0.4$}

\end{axis}
\end{tikzpicture}
\caption{Per-layer retention ratio under AL-AM allocation for
LLaMA-2-7B at four pruning ratios. Values above the dashed
line ($1.0$) indicate layers where more parameters are
retained relative to a uniform baseline. At mild pruning
($p=0.1$), the allocation is nearly uniform; as the ratio
increases, the U-shaped pattern intensifies, with first and
last layers increasingly preserved while middle layers
(12--22) are pruned most aggressively.}
\label{fig:layer-sparsity}
\end{figure}

Figure~\ref{fig:layer-sparsity} visualizes the per-layer
retention ratio under AL-AM allocation for LLaMA-2-7B. At
mild pruning ($p=0.1$), the allocation is nearly uniform. As
the ratio increases, a U-shaped pattern emerges: the first
and last layers are increasingly preserved while middle
layers are pruned most aggressively.

\section{Qualitative Generation Examples}
\label{app:qualitative}

Table~\ref{tab:qualitative} presents generated text from the dense LLaMA-2-7B baseline and TriSP-pruned models (20\% and 30\% with LoRA recovery) using greedy decoding with a repetition penalty. Our experiments demonstrate that the pruned LLaMA-2-7B models with 5.43B and 4.79B parameters, obtained through our TriSP pruning approach, are highly effective in retaining general knowledge and producing fluent, coherent text.

\end{document}